\documentclass{article}

\usepackage [latin1]{inputenc}
\usepackage{natbib}
\usepackage{mathtools}
\usepackage{multirow}
\usepackage{hhline}
\usepackage{graphics}
\usepackage{epsfig}
\usepackage{amssymb}
\usepackage{amsmath}
\usepackage{tikz} 
\usetikzlibrary{calc}
\usetikzlibrary{decorations.pathreplacing}
\usetikzlibrary{angles,quotes}

\title{A Dataset-Level Geometric Framework for Ensemble Classifiers}

\author{Shengli Wu \\
       School of Computing, Ulster University, Belfast, UK\\
       Weimin Ding \\
       School of Computer Science, Jiangsu University, Zhenjiang, China}

\begin{document}

\maketitle

\begin{abstract}

Ensemble classifiers have been investigated by many in the artificial intelligence 
and machine learning community. Majority voting and weighted majority voting 
are two commonly used combination schemes in ensemble learning.
However, understanding of them is 
incomplete at best, with some properties even misunderstood.  
In this paper, we present a group of properties of these two schemes formally 
under a dataset-level geometric framework. Two key factors, every component base 
classifier's performance and dissimilarity between each pair of component classifiers 
are evaluated by the same metric - the Euclidean distance. Consequently, 
ensembling becomes a deterministic problem and the performance of an ensemble 
can be calculated directly by a formula. We prove several theorems of interest 
and explain their implications for ensembles. In particular, we compare and contrast 
the effect of the number of component classifiers on these two types of ensemble schemes. 
Empirical investigation is also conducted to verify the theoretical results 
when other metrics such as accuracy are used. We believe that the results 
from this paper are very useful for us to understand the fundamental properties 
of these two combination schemes and the principles of ensemble classifiers in general. 
The results are also helpful for us to investigate some issues in ensemble classifiers, 
such as ensemble performance prediction, selecting a small number of base classifiers 
to obtain efficient and effective ensembles.

\end{abstract}


\section{Introduction}

In the last three decades, ensemble learning has been investigated by many researchers. 
This technique has seen use in diverse tasks such as classification, regression, 
and clustering among others. Research has been conducted at multiple levels: 
from feature selection, component classifier generation and selection, 
to the ensemble model  \citep{JurekBWN14,DongYCSM20}. Some of the ensemble approaches 
have been very successful in international machine learning competitions such as Kaggle, 
KDD-Cups, etc. and the technologies have been used in various application areas  \citep{OzaT08a}.

Although some ensemble models (such as stacking, bagging, random forest,
AdaBoost, gradient boosting machines, deep neural network-based models, and others)
are more complicated, two relatively simple combination schemes,
majority voting and weighted majority voting, have been
used widely for the ensemble model \citep{Sagi&Rokach18}.
Even in those more complicated models, majority voting and weighted majority voting
are still used frequently at intermediate or final combination stages.


How many component classifiers to use and how to select a subset from a large 
group for an ensemble are two related questions. 
\cite{Zhou&Wu02} investigated the impact of the number of component classifiers on
ensemble performance. It is found that the number of component classifiers
is not always a positive factor for improving performance when majority voting is used
for combination. Later their finding is often referred to as the 
``many-could-be-better-than-all'' theorem.

However, this finding is not echoed by many others. 
More empirical evidence indicates that the size of an ensemble 
has a positive impact on performance \citep{Hernandez-LobatoMS13}.
To balance ensemble performance and efficiency, many papers investigate
how to achieve best possible performance by combining a fixed or a small number of classifiers
selected from a large number of candidates 
\citep{LatinneDD01,OshiroPB12,XiaoHJL10,DiasW14,BhardwajBS16,YkhlefB17,ZhuNNJCL19}.

Possibly inspired by \cite{Zhou&Wu02}'s work, 
\cite{BonabC19} asserted that for weighted majority voting: 
the ideal condition for the ensemble to achieve 
maximum accuracy is for the number of component classifiers to equal
the number of output classes. However, their theoretical analysis is limited 
by the strength of assumptions used, which experimental results do not always support.

Diversity among component classifiers is a factor that may influence ensemble performance \citep{Bi12,JainLM020,ZhangJS020}. 
However, there is no generally accepted definition of diversity \citep{KunchevaW03}.
Many measures, such as Yule's Q statistics, the correlation coefficient, the disagreement
measure, F-score, the double fault measure,  Kohavi-Wolpert's variance,
Kuncheva's entropy, etc., have been proposed and investigated \citep{TangSY06,VisentiniSF16}.
Moreover, many diversity measures such as the F-score, the disagreement
measure, the double fault measure, etc. are or related to performance measures.
This explains why the effect of diversity on ensemble performance is unclear \citep{TangSY06,Bi12}.

In short, although ensemble learning has garnered considerable attention from researchers, 
much of the literature comprise methodological and empirical studies, while the theory is 
underrepresented. Many fundamental questions remain unclear. We list some of these below:

\begin{enumerate}

\item What is the difference between majority voting and weighted majority voting?

\item When should we use majority voting rather than weighted majority voting, or vice versa?   

\item There are a lot of weighting assignment methods for weighted majority voting, which one is the best?

\item How does the number of component classifiers affect ensemble performance?

\item How does each component classifier affect ensemble performance?

\item How does performance of component classifiers affect ensemble performance?

\item How does diversity of component classifiers affect ensemble performance?

\end{enumerate}

The goal of this paper is to set up a geometric framework for ensemble classifiers, 
thereby enabling us to clearly answer the above questions. 
In this framework, the result from each base classifier is represented as a point 
in a multi-dimensional space. We can use the same measure - the Euclidean distance - for 
both performance and diversity. Ensemble learning becomes a deterministic problem. 
Many interesting theorems about the relation of component classifiers and ensemble 
classifiers can be proven. Armed with this framework, we discover and present some 
useful features of majority voting and weighted majority voting.

The rest of this paper is organized as follows: Section 2 presents some related work.
In Section 3, we present the geometric framework for ensemble classifiers. 
Some characteristics of majority voting and weighted majority voting are presented 
and compared. Discussion regarding the questions raised previously is given in Section 4, 
supported by theory. In Section 5, we present some empirical investigation results 
to confirm our findings in Section 3. Finally Section 6 is the conclusion.

\section{Related work}

Majority voting and weighted majority voting are commonly used in many
ensemble models. Their features and some related tasks have been investigated 
by many researchers.

\cite{Zhou&Wu02} investigated majority voting 
theoretically and empirically through an ensemble of a group of neural networks.
It is found that selecting a subset
may be able to achieve better performance than combining all available classifiers. 
Note that in both their theoretical and empirical study, 
weighted majority voting is mentioned, but not considered.

\cite{FumeraRS08} set up a probabilistic framework to analyse 
the bagging misclassification rate when the ensemble size increases. 
Majority voting is used for combination.
Both their theoretical analysis and empirical investigation
show that bagging misclassification rate decreases when more and more component
classifiers are combined. This is somewhat inconsistent with the 
finding of \cite{Zhou&Wu02}.

\cite{LatinneDD01} used the McNemar test to decide if two sets of component classifiers
are significantly different. More and more classifiers are added to the pool
and the McNemar test is carried out repeatedly,
the process stops if no significant difference can be observed. 

\cite{OshiroPB12} raised the problem of how many component classifiers (trees) should be used 
for an ensemble (random forest). Experimenting with 29 real datasets in the biomedical domain, 
they observed that higher accuracy is achievable when more 
trees are combined by majority voting. However, when the number of trees is relatively large 
(say, over 32 or 64), the improvement is no longer significant. Therefore, a good choice 
is to consider both ensemble accuracy and computing cost.
This issue is also addressed in \cite{Hernandez-LobatoMS13,AdnanI16,ProbstB17}.

Many researchers find that if more base classifiers are combined,
then the ensemble is able to achieve better performance. 
Although it is possible to achieve better performance by fusing more base classifiers, 
it costs more. Quite a few papers investigated the ensemble pruning problem that aims at 
increasing efficiency by reducing the number of base classifiers, without losing much 
performance at the same time. Various kinds of methods have been proposed. See 
\cite{XiaoHJL10,DiasW14,BhardwajBS16,YkhlefB17,ZhuNNJCL19} for some of them.

How to assign proper weights for weighted majority vote has been investigated by
quite a few researchers.
Some methods may consider different aspects of base classifiers for the weights: 
performance-based in \cite{Opitz&Shavlik96,Wozniak08}, 
a model probability-based in \cite{Duan&Ajami07},
prediction confidence-based in \cite{Schapire&Singer99},
and Matthews correlation coefficient-based in \cite{HaqueNBM16}.
Some methods may have different goals for weight optimisation:
minimizing classification error in \cite{Kuncheva&Diez14,Mao&Jiao15, Bashir&Qamar15},
reducing variance while preserving given accuracy in \cite{Derbeko&El-Yaniv02},
minimizing Euclidean distance in \cite{BonabC18}. 
Different search methods are also used: 
A linear programming-based method is presented in \cite{ZhangZ11},
\cite{GeorgiouMT06} applied a game theory to weight assignment,
\cite{Liu&Cao14} applied the genetic algorithm to weight assignment.
Dynamic adjustment of weights is investigated in \cite{Valdovinos&S09}.

\cite{Wu&Crestani15} proposed a geometric framework for data fusion in information retrieval.
In this query-level framework, for a certain query each component retrieval system assigns 
scores to all the documents in the collection, which indicates their relevance to the query.
Also predefined values are given to documents that are relevant or irrelevant to the query, which 
are ideal scores for the documents involved.
Therefore, the scores from each component retrieval system can be regarded as a point in a multiple
dimensional space, and the ideal scores form an ideal point.
Under this framework, performance and dissimilarity can be measured by 
the same metric - the Euclidean distance. Both majority voting and 
weighted majority voting can be explained and calculated in the geometric space.
However, the framework is set up at the query level, which is equivalent to the instance level
in ensemble classifiers.

\cite{BonabC18,BonabC19} adapted the model in \cite{Wu&Crestani15} to ensemble classifiers.
Some properties of majority voting and weighted majority voting are presented. 
However, as in \cite{Wu&Crestani15}, their framework is at the instance level.
This means that the theorems hold for each individual instance, but it is  
unclear if they remain true for multiple instances collectively. 
The latter is a more important and realistic situation we should consider.
As we know, a training dataset or a test dataset usually comprises a group of instances. 
It is desirable to know the collective properties of 
an ensemble classifier over all the instances, rather than that of any individual instance. 
This generalization is the major goal of this paper. 

To do this, we first define a dataset-level framework 
and then go about proving a number of useful theorems.

\section{The geometric framework}

\begin{table}
\caption{Notation used in this paper}
\begin{center}
\begin{tabular}{|ll|} \hline

Symbols & \qquad \hspace{1in} Meaning \\

\hline

$C$ & Centroid of a group of points in $X$ \\

$CL$ & A set of classes \\

$cl_j$ & One of the classes in $CL$ \\

$CF$ & A set of component classifiers \\

$cf_k$ & One of the component classifiers in $CF$ \\

$DT$ & A dataset has components $CF$, $CL$, $S$, $T$, and $O$ \\

$ed(S^i,S^j)$ & the Euclidean distance between two points $(S^i,S^j)$ in $X$ \\

$F$ & Fused point by linear combination of a group of points in $X$ \\

$m$ & Number of component classifiers in $CF$ \\

$n$ & Number of instances in $T$ \\

$O$ & All the real labels for $T$ constitute the ideal point in $X$ \\

$o_{ij}$ & One of the elements of $O$ \\

$o_{l}$ & Another form of $o_{ij}$, in which $l=(i-1)*n+j$ \\

$p$ & Number of classes in $CL$ \\  

$\mathcal{S}$ & Scores given by $CF$ members relating to $CL$, a set of points in $X$ \\  

$S^k$ &  Scores given by $cf_k$ for $T$ relating to $CL$, a point in $X$ \\

$s_{ij}^k$ & A score given by $cf_k$ for $t_i$ relating to $c_j$ \\

$s_{l}^k$ & Another form of  $s_{ij}^k$, in which $l=(i-1)*n+j$ \\

$T$ & A set of instances \\

$t_i$  & One of the instances in $T$  \\

$w^k$  & Weight assigned to component classifier $cf_k$ \\

$X$ &  a $n*p$-dimensional space relating to $DT$ \\  \hline

\end{tabular}
\end{center}
\end{table}

In this section, we introduce the dataset-level geometric framework.

Suppose for a machine learning problem we have $p$ classes
$CL$=\{$cl_1$,$cl_2$,$\cdots$,$cl_p$\},
the ensemble has $m$ component base classifiers
$CF$=\{$cf_1$,$cf_2$,$\cdots$,$cf_m$\}, 
and the dataset $DT$ has $n$ instances $T$=\{$t_1$,$t_2$,$\cdots$,$t_n$\}.
For every instance $t_i$ and every class $cl_j$, each base classifier $cf_k$
provides a score $s_{ij}^{k}$, which indicates the estimated probability score that 
$t_i$ is an instance of class $cl_k$ given by $cf_j$. 
$s_{ij}^{k} \in [0,1]$. Each instance $t_i$ has a real label 
for each class $cl_k$, which is 0 or 1. 
We may set up a $n*p$-dimensional space for the above problem.
There are $m$ points $\{S^1,S^2,\cdots,S^m\}$, each represents the scores given by a specific classifier to
all the instances in the dataset for all the classes. Point $S^k$ is:

\begin{eqnarray}
\nonumber 
S^k&=&\{(s_{11}^k,s_{12}^k,\cdots,s_{1p}^k),(s_{21}^k,s_{22}^k,\cdots,s_{2p}^k),\cdots(s_{n1}^k,s_{n2}^k,\cdots,s_{np}^k)\} \\ \nonumber 
&=&\{s_{1}^k,s_{2}^k,\cdots,s_{n*p}^k\} \\ \nonumber 
\end{eqnarray}

As such, we can always organize all the scores first by instances and then by classes. 
Thus a two-dimensional array with $n$ elements in one dimension and $p$ in the other
is transformed to a list of $n*p$ elements:
$s_{ij}^{k}$ becomes $s_{(i-1)*n+j}^k$
The ideal point is also represented in the same style.

\begin{eqnarray}
\nonumber 
O&=&\{(o_{11},o_{12},\cdots,o_{1p}),(o_{21},o_{22},\cdots,o_{2p}),\cdots(o_{n1},o_{n2},\cdots,o_{np})\} \\ \nonumber 
&=&\{o_{1},o_{2},\cdots,o_{n*p}\} \\ \nonumber 
\end{eqnarray}

\noindent which indicates the real labels of every instance in the dataset to each of the classes involved:
1 for a true label and 0 for a false label.
The notation used in this paper is summarized in Table 1.

This framework is a generalization of the one presented in \cite{BonabC18,BonabC19}.
If the dataset only has one instance, then the above framework is the same as the one
in \cite{BonabC18,BonabC19}. This framework is suitable for soft voting \citep{CaoKWLLK15},
in which each component classifier provides probability scores for every instance 
relating to each class.
If no such probability scores are provided (hard voting), then it is still possible to apply 
it to the geometric framework if we transform estimated class labels into proper scores.
The geometric framework is applicable to both single-label and multi-label classification problems.

\noindent \textbf{Example 1.} A dataset includes two instances $t_1$ and $t_2$,
and three classes $cl_1$, $cl_2$, and $cl_3$. 
$t_1$ is an instance of $cl_2$ but not the other two,
and $t_2$ is an instance of both class $cl_1$ and $cl_3$ but not $cl_2$. 
The scores got from the three base component classifiers  
$cf_1$, $cf_2$, and $cf_3$ are as follows:

\begin{center}

\smallskip
\begin{tabular}{|c|r|r|r|r|} \hline

Instance  & Classifier  & Class $cl_1$ & Class $cl_2$ & Class $cl_3$ \\ \hline
\multirow{3}{*}{$t_1$}  & $cf_1$  & $s_{11}^1=0.5$ &  $s_{12}^1=0.6$  & $s_{13}^1=0.3$ \\ \hhline{~----}
  & $cf_2$ & $s_{11}^2=0.4$ & $s_{12}^2=0.7$ & $s_{13}^2=0.2$ \\ \hhline{~----}            
  & $cf_3$ & $s_{11}^3=0.6$ & $s_{12}^3=0.8$ & $s_{13}^3=0.4$ \\ \hhline{~----}            
 & Real label & $o_{11}=0$ & $o_{12}=1$ & $o_{13}=0$ \\ \hline                 
\multirow{3}{*}{$t_2$}  & $cf_1$  & $s_{21}^1=0.7$ &  $s_{22}^1=0.3$  & $s_{23}^1=0.9$ \\ \hhline{~----}
 & $cf_2$ & $s_{21}^2=0.3$ & $s_{22}^2=0.6$ & $s_{23}^2=0.7$ \\ \hhline{~----}               
 & $cf_3$ & $s_{21}^3=0.2$ & $s_{22}^3=0.6$ & $s_{23}^3=0.8$ \\ \hhline{~----}               
 & Real label & $o_{21}=1$ & $o_{22}=0$ & $o_{23}=1$ \\ \hline

\end{tabular}
\end{center}

In a 6-dimensional geometric space, we may set up four points to represent 
this scenario:
$S^1=\{0.5,0.6,0.3,0.7,0.3,0.9\}$, $S^2=\{0.4,0.7,0.2,0.3,0.6,0.7\}$,
$S^3=\{0.6,0.8,0.4,0.2,0.6,0.8\}$,
and $O=\{0,1,0,1,0,1\}$.   \hfill $\Box$

In the following we use point and component result interchangeably
if no confusion will be caused. 
We can calculate the Euclidean distance of two points $S^{u}$ and $S^{v}$ 

\begin{equation}
ed(S^u,S^v)=\sqrt{{\sum_{i=1}^{n}\sum_{j=1}^{p}{(s_{ij}^u-s_{ij}^v)}^2}}=\sqrt{\sum_{l=1}^{n*p}{(s_{l}^u-s_{l}^v)}^2}
\end{equation}

We may use the Euclidean distance to evaluate the performance of 
classifier $cf_u$ over $n$ instances and $p$ classes.

\begin{equation}
ed(S^u,O)=\sqrt{{\sum_{i=1}^{n}\sum_{j=1}^{p}{(s_{ij}^u-o_{ij})}^2}}=\sqrt{\sum_{l=1}^{n*p}{(s_l^u-o_l)}^2}
\end{equation}

It is an advantage of the geometric framework to evaluate both performance of a component result 
and dissimilarity of two component results by using the same
metric. They will be referred to as performance distance  and dissimilarity distance later in this paper.


\noindent \textbf{Definition 1 (Performance distribution).}
In a geometric space $X$, there are $m$ points $\mathcal{S}$=$\{S^1,S^2,\cdots,S^m\}$ ($m \ge 1$). 
$O$ is the ideal point. Performance distribution of these $m$ points in $\mathcal{S}$
is defined as $ed(S^i,O)$ for ($1 \leq i \leq m$).

\noindent \textbf{Definition 2 (Dissimilarity distribution).}
In a geometric space $X$, there are $m$ points $\mathcal{S}$=$\{S^1,S^2,\cdots,S^m\}$ ($m \ge 1$). 
Dissimilarity distribution of these $m$ points in $\mathcal{S}$
is $ed(S^i,S^j)$ for ($1 \leq i \leq m$, $1 \leq j \leq m$, $i \neq j$).

From their definitions, we can see that performance distribution 
and dissimilarity distribution are two completely different aspects of $\mathcal{S}$, 
and not related to each other. Their independence is a good thing 
for us to investigate the properties
of ensembles, especially the effect of each on ensemble performance.

\subsection{Majority voting}

In a $n*p$-dimensional space, there are $m$ points $\mathcal{S}$=\{$S^1$,$S^2$,$\cdots$,$S^m$\}
($m \ge 2$). Combining them by majority voting can be understood to be finding the centroid of these $m$ points.
It is referred to as the centroid-based fusion method in \cite{Wu&Crestani15}.

\noindent \textbf{Theorem 1.}
In a geometric space $X$, there are $m$ points $\mathcal{S}$=$\{S^1,S^2,\cdots,S^m\}$ ($m \ge 2$). 
$C$ is the centroid of these $m$ points and $O$ is the ideal point.
The distance between $C$ and $O$ is no longer than the average distance between 
each of the $m$ points and $O$:

\begin{equation}
\label{eq:theorem1}
\sqrt{\sum_{l=1}^{n*p}{(c_l-o_l)^2}} \leq \frac{1}{m}\sum_{k=1}^{m}\sqrt{\sum_{l=1}^{n*p}{(s_{l}^{k}-o_l)^2}}
\end{equation}

Proof:  Replace $C$ with its definition $c_l=\frac{1}{m}\sum_{k=1}^{m}{s_l^k}$ for ($1 \leq l \leq n*p$) in Equation \ref{eq:theorem1}
and move $\frac{1}{m}$ on the right side to the left as $m$, we get 

\begin{equation}
\label{eq:theorem1-2}
m\sqrt{\sum_{l=1}^{n*p}{(\frac{1}{m}\sum_{k=1}^{m}{s_l^k}-o_l)^2}} \leq \sum_{k=1}^{m}\sqrt{\sum_{l=1}^{n*p}{(s_{l}^{k}-o_l)^2}}
\end{equation}

The Minkowski Sum Inequality \citep{Minkowski} is

\[ {(\sum_{l=1}^{n*p}{{(a_l+b_l)}^q})}^{q^{-1}} \leq {\sum_{l=1}^{n*p}{a_l^q}}^{q^{-1}}+{\sum_{l=1}^{n*p}{b_l^q}}^{q^{-1}} \]

\noindent where $q>1$ and $a_l$, $b_l$ $>$ 0. For our question, we let $q=2$ and 
$a_l=s_l^k-o_l$. Then we have:

\[ \sum_{k=1}^m{\sqrt{\sum_{l=1}^{n*p}{{(s_l^k-o_l)}^2}}} = \sum_{k=1}^m\sqrt{\sum_{l=1}^{n*p}{(a_l)}^2}  \]

\[ \geq \sqrt{\sum_{l=1}^{n*p}{{(a_1+a_2)}^2}} + \sum_{k=3}^m\sqrt{\sum_{l=1}^{n*p}{(a_l)}^2} \]

\[ \geq ... \geq \sqrt{\sum_{l=1}^{n*p}{{(a_1+a_2+...+a_m)}^2}} = \sqrt{\sum_{l=1}^{n*p}{(\sum_{k=1}^m{a_k})^2}} \]

Now notice the left side of Inequation \ref{eq:theorem1-2}

\[ m\sqrt{\sum_{l=1}^{n*p}{{(\frac1m\sum_{k=1}^m{s_l^k}-o_l)^2}}} = \sqrt{\sum_{l=1}^{n*p}{(\sum_{k=1}^{m}{s_l^k}-m*o_l)^2}} = \sqrt{\sum_{l=1}^{n*p}({\sum_{k=1}^m{a_k}})^2} \ \ \ \ \Box \] 


This theorem tells us, if we take all the instances
together and use the Euclidean distance
as the performance metric, then the performance of the ensemble by majority voting 
is at least as good as the average performance of all component
classifiers involved. This theorem confirms the observation 
by many researchers that the results from the ensemble are stable and 
good.

\noindent \textbf{Example 2.} Consider the points in Example 1. 
$C=\{0.5,0.7,0.3,0.4,0.5,0.8\}$, $ed(S^1,O)=0.83$, 
$ed(S^2,O)=1.11$, $ed(S^3,O)=1.26$,
$ed(C,O)=1.04$,  $(ed(S^1,O)+ed(S^2,O)+ed(S^3,O))/3=1.06$.
$ed(C,O)$ is slightly smaller than the average of $ed(S^1,O)$, $ed(S^2,O)$, and $ed(S^3,O)$. \hfill $\Box$

\noindent \textbf{Theorem 2.} 
In a space $X$, suppose that $\mathcal{S}$=\{$S^1$, $S^2$,$\cdots$, $S^m$\} and $O$ are known points.
$C$ is the centroid of $S^1$, $S^2$,$\cdots$, $S^m$.
The distance between $C$ and $O$ can be represented as

\begin{equation}
\label{eq:theorem2}
ed(C,O)=\frac1m\sqrt{m\sum_{i=1}^m{ed(S^i,O)}^2-\sum_{i=1}^{m-1}\sum_{j=i+1}^m{ed(S^i,S^j)}^2}
\end{equation}

Proof. Assume that $O$=(0,...,0), which can always be done by coordinate 
transformation. According to its definition 
\(  C=(\frac1m\sum_{i=1}^m{s_1^i},..., \frac1m\sum_{i=1}^m{s_{n*p}^i}) \), we have

\begin{equation} \label{eq:T2-1}
\begin{split}
 ed(C,O)&=\sqrt{{(\frac1m\sum_{i=1}^m{s_1^i})^2}+...+{(\frac1m\sum_{i=1}^m{s_{n*p}^i})^2}} \\
 &=\frac{1}{m}\sqrt{\sum_{i=1}^m\sum_{k=1}^{n*p}{(s_k^i)}^2+2\sum_{k=1}^{n*p}\sum_{i=1}^{m-1}\sum_{j=i+1}^m{s_k^i*s_k^j}}
\end{split}
\end{equation}

Note that the distance between $S^i$ and $S^j$ is
\( ed(S^i,S^j)=\sqrt{\sum_{k=1}^{n*p}{(s_k^i-s_k^j)}^2} \)

or \( {ed(S^i,S^j)}^2=\sum_{k=1}^{n*p}{(s_k^i)^2}+\sum_{k=1}^{n*p}{(s_k^j)^2}-2\sum_{k=1}^{n*p}{s_k^i*s_k^j} \)

Because \( {ed(S^i,O)}^2=\sum_{k=1}^{n*p}{(s_k^i)}^2 \)
and \( {ed(S^j,O)}^2=\sum_{k=1}^{n*p}{(s_k^j)}^2 \), we get

\[ 2\sum_{k=1}^{n*p}{s_k^i*s_k^j}={ed(S^i,O)}^2+{ed(S^j,O)}^2-{ed(S^i,S^j)}^2 \]

Considering all possible pairs of points $S^i$ and $S^j$ and we get

\[  2\sum_{k=1}^{n*p}\sum_{i=1}^{m-1}\sum_{j=i+1}^m{s_k^i*s_k^j}=(m-1)\sum_{i=1}^m{ed(S^i,O)}^2-\sum_{i=1}^{m-1}\sum_{j=i+1}^m{ed(S^i,S^j)}^2  \]

In Equation \ref{eq:T2-1}, we use the right side of the above equation to replace

\[  2\sum_{k=1}^{n*p}\sum_{i=1}^{m-1}\sum_{j=i+1}^m{s_k^i*s_k^j} \]

Also note that \( \sum_{i=1}^m\sum_{k=1}^{n*p}({s_k^i})^2=\sum_{i=1}^m{ed(S^i,O)}^2 \),
we obtain

\[ ed(C,O)=\frac1m\sqrt{m\sum_{i=1}^m{ed(S^i,O)}^2-\sum_{i=1}^{m-1}\sum_{j=i+1}^m{ed(S^i,S^j)}^2} \qquad \qquad \qquad \Box\]

This theorem tells us that the ensemble performance is completely decided by
$ed(S^i,O)$ (for $1 \leq i \leq m$) and $ed(S^i, S^j)$ 
(for $1 \leq i \leq m$, $1 \leq j \leq m$, $i \ne j$).
The impact of both performance of component classifiers and dissimilarity of all
pairs of component classifiers on ensemble performance can be seen clearly.
According to Equation 2, 
in order to minimize $ed(C,O)$, we need to minimize $\sum_{i=1}^{m}{ed(S^i,O)^2}$
and maximize $\sum_{i=1}^{m-1}\sum_{j=i+1}^{m}{ed(S^i,S^j)^2}$ at the same time.
Therefore, for performance distribution, both average and variance affect ensemble performance.
Lower average and lower variance lead to better performance. 
For dissimilarity distribution, both average and variance also affect ensemble performance. 
Higher average and higher variance lead to better performance. 

\noindent \textbf{Example 3.} Assume that $\mathcal{S}^1$=\{$S^1$, $S^2$\},
$\mathcal{S}^2$=\{$S^3$, $S^4$\}, $ed(S^1,S^2)=ed(S^3,S^4)$, $ed(S^1,O)$=0.5, $ed(S^2,O)$=0.5,
$ed(S^3,O)$=0.4, $ed(S^4,O)$=0.6. 
	The centroid of $S^1$ and $S^2$ is $C^1$, and the centroid of $S^3$ and $S^4$ is $C^2$.
We have $cd(C^1,O)$ $<$ $cd(C^2,O)$.

In this example, because dissimilarity distance is the same for both $\mathcal{S}^1$ and $\mathcal{S}^2$, 
we only need to consider performance distribution.
The average of $ed(S^1,O)$ and $ed(S^2,O)$ is 0.5, and  
the average of $ed(S^3,O)$ and $ed(S^4,O)$ is also 0.5. 
The variance of $\mathcal{S}^1$ is smaller than that of $\mathcal{S}^2$.
$ed(S^1,O)^2+ed(S^2,O)^2$=0.50, 
$ed(S^3,O)^2+ed(S^4,O)^2$=0.52.
Because $ed(S^1,S^2)=ed(S^3,S^4)$, we get $ed(C^1,O)$ $<$ $ed(C^2,O)$. \hfill $\Box$

\noindent \textbf{Example 4.} In the figure below, there are six points $S^i$ $(1 \leq i \leq 6)$.
Among these points, $S^1$ is the closest to $O$.
It is followed by $S^2$ and $S^3$, which are equally distant to $O$.
Finally, $S^4$, $S^5$ and $S^6$ are equally distant to $O$ and they are 
all further from $O$ than the other three points. Now we try to work out a subset of 
these to maximize performance.

\begin{center}
\begin{tikzpicture}

\draw [black,dashed]  (0,0) circle [radius=1];;
\draw [black,dashed]  (0,0) circle [radius=1.5];;
\draw [black,dashed]  (0,0) circle [radius=2];;

\draw [fill] (0,0) circle [radius=0.06];
\node [left] at (0,0) {$O$};

\draw [fill] (0,-1) circle [radius=0.06];
\node [below] at (0,-1) {$S^1$};



\draw [fill] (-1.42,0.5) circle [radius=0.06];
\node [left] at (-1.42,0.5) {$S^2$};

\draw [fill] (1.42,0.5) circle [radius=0.06];
\node [right] at (1.42,0.5) {$S^3$};

\draw [fill] (-1.732,-1) circle [radius=0.06];
\node [left] at (-1.732,-1) {$S^4$};

\draw [fill] (1.732,-1) circle [radius=0.06];
\node [right] at (1.732,-1) {$S^5$};

\draw [fill] (0,2) circle [radius=0.06];
\node [above] at (0,2) {$S^6$};

\draw[dashed] (-1.42,0.5) -- (1.42,0.5);
\draw[dashed] (0,0) -- (0,-1);

\draw [fill] (0,0.5) circle [radius=0.06];
\node [right] at (0,0.5) {$C^2$};

\draw[dashed] (-1.732,-1) -- (0,0);
\draw[dashed] (1.732,-1) -- (0,0);
\draw[dashed] (0,2) -- (0,0);

\end{tikzpicture}
\end{center}

Now, if we select one point only, then $S^1$ is the best option;
if we select two points, then combining $S^2$ and $S^3$ is the best option, 
in this instance the centroid would be $C^2$;
if we select three points, then combining $S^4$, $S^5$ and $S^6$ is the best option,
this gives the centroid $O$.
Fusing all six points is also a good option, but it may not be as good as fusing $S^4$, 
$S^5$ and $S^6$. The ``many-could-be-better-than-all'' theorem seems reasonable \citep{Zhou&Wu02} 
in this case. Because the centroid of a group of points is decided by the positions of 
all the points collectively, each of which has an equal weight, removing or 
adding even a single point may change the position of the centroid
of a group of points significantly. It also indicates that it is not an easy task 
to find the best subset from a large group of base classifiers.
As a matter of fact, it is an NP-hard problem, as our next theorem proves. \hfill $\Box$

\noindent \textbf{Theorem 3.} 
$\mathcal{S}$ = \{$S_1$, $S_2$,$\cdots$, $S_m$\} for $(m \ge 3)$ is a group of points and $O$ is an ideal point. 
For a given number $m'$ ($2 \leq m'<m$), the problem is to find a subset of $m'$ points 
from $\mathcal{S}$ to minimize the distance of the centroid of 
these $m'$ points to the ideal point $O$. The above question is NP-hard.

Proof: First let us have a look at the maximum diversity problem (MDP), which is a known NP-hard 
problem \citep{WangHGL14}. The MDP is to identify a subset $\mathcal{E'}$ of $m'$ elements from 
a set $\mathcal{E}$ of $m$ elements,
such that the sum of the pairwise distance between the elements in the subset is maximized.
More precisely, let $\mathcal{E}$ = \{$e_1$, $e_2$,$\cdots$, $e_m$\} be a group of elements,
and $d_{ij}$ be the distance between elements $e_i$ and $e_j$. The objective of the MDP can be 
formulated as:

Maximize 

\begin{equation}
\label{mdp}
f(x)=\frac{1}{2}\sum_{i=1}^{m}\sum_{j=1}^{m}d_{ij}*x_{i}*x_{j}
\end{equation}

subject to \(\sum_{i=1}^{m}x_{i}=m', x_{i} \in \{0,1\}, i=1,\cdots, m\)

where each $x_i$ is a binary (zero-one) variable indicating whether an element
$e_i \in \mathcal{E}$ is selected to be a member of $\mathcal{E'}$.

On the other hand, according to Theorem 2, our problem may be written as minimizing ${ed(C,O)}^2$.
Here $C$ is the centroid of $m'$ selected points.

\[ {ed(C,O)}^2=\frac{1}{m'}\sum_{i=1}^m{ed(S^i,O)}^2*x_i-\frac{1}{m'*m'}\sum_{i=1}^{m}\sum_{j=1}^{m}{ed(S^i,S^j)}^2*x_i*x_j \]   

subject to \(\sum_{i=1}^{m}x_{i}=m', x_{i} \in \{0,1\}, i=1,\cdots, m\)

where each $x_i$ is also a binary (zero-one) variable indicating whether a point
$S^i \in \mathcal{S}$ is selected to be a member of $\mathcal{S'}$.

If we assume that $ed(S^i,O)$ equals to each other for all  $i=1,\cdots, m$, then 
$\frac{1}{m'}\sum_{i=1}^m{ed(S^i,O)}^2*x_i$ becomes a constant and
minimizing ${ed(C,O)}^2$ equals to maximizing $div(C,O)^2$

\[{div(C,O)}^2=\frac{1}{m'*m'}\sum_{i=1}^{m}\sum_{j=1}^{m}{ed(S^i,S^j)}^2*x_i*x_j\]   

Let $g(x)$=$div(C,O)^2$ and $d_{ij}=\frac{2}{m'*m'}ed(S^i,S^j)^2$, 
then the above equation can be rewritten as

\begin{equation}
\label{div}
g(x)=\frac{1}{2}\sum_{i=1}^{m}\sum_{j=1}^{m}d_{ij}*x_i*x_j   
\end{equation}

Comparing Equation \ref{mdp} and \ref{div}, we can see that 
a simplified version of our problem is 
an MDP problem. Therefore, our problem is 
an NP-hard problem.  \hfill $\Box$

Theorem 3 tells us: for a given number of classifiers, 
choosing a subset for best ensemble performance by majority voting
is an NP-hard problem.

Although the ``many-could-be-better-than-all'' theorem may be applicable to some cases, 
it does not tell the whole story. Now let us have a look at how the size of the ensemble 
impacts its performance. Because the performance of an ensemble is affected by 
a few different factors, we need to find a way of separating this from other factors.

\noindent \textbf{Theorem 4.}
In a space $X$, $\mathcal{S}$ = $\{S^1,S^2,\cdots,S^m\}$ and $O$ are known points.
$C$ is the centroid of $S^1,S^2,\cdots,S^m$, 
$C^1$ is the centroid of $m-1$ points $S^2,S^3,\cdots,S^m$, 
$C^2$ is the centroid of $m-1$ points $S^1,S^3,\cdots,S^m$,$\cdots$, 
$C^m$ is the centroid of $m-1$ points $S^1,S^2,\cdots,S^{m-1}$. We have 

\begin{equation}
\label{eq:theorem4}
ed(C,O)\leq\frac{1}{m}\sum_{i=1}^{m}{ed(C^i,O)}
\end{equation}  

Proof: According to the definition of $C^1$, $C^2$, $\cdots$, $C^m$, 
$C$ is the centroid of these $m$ points. 
The theorem can be proven by applying Theorem 1. \hfill $\Box$

Theorem 4 can be used repeatedly to 
prove more general situations in which 
a subset includes $m-2$,$m-3$,$\cdots$,or 2 points.  
This demonstrates that the number of component results has a positive effect 
on ensemble performance.

\noindent \textbf{Example 5.}
In a space $X$, $\{S^1,S^2,S^3,S^4\}$ and $O$ are known points. 
There are four different combinations of three points and six combinations of two points.
We use $C^{1234}$ to represent the centroid of all $4$ points.
Similarly, $C^{23}$ represents the centroid of $S^2$ and $S^3$, and so on.
Applying Theorem 4 repeatedly, we have

\begin{eqnarray}
\nonumber 
ed(C^{1234},O)&\leq&\frac{1}{4}[ed(C^{123},O)+ed(C^{124},O)+ed(C^{134},O)+ed(C^{234},O)]   \\ \nonumber   
              &\leq&\frac{1}{6}[ed(C^{12},O)+ed(C^{13},O)+ ed(C^{14},O) \\ \nonumber  
              & & +ed(C^{23},O)+ed(C^{24},O)+ed(C^{34},O)] \\ \nonumber 
              &\leq&\frac{1}{4}[ed(S^{1},O)+ed(S^{2},O)+ed(S^{3},O)+ed(S^{4},O)] \quad \Box \\ \nonumber 
\end{eqnarray}

Although Theorem 4 shows that the number of component classifiers has a positive effect on
ensemble performance, it is not clear how significant the effect is. 
Next let us look at this matter quantitatively.
In order to focus on the number of component classifiers, 
we make a few simplifying assumptions.
Suppose that $S^1$, $S^2$,$\cdots$, $S^m$ and $O$ are known points 
in space $X$. $ed$($S^i$, $O$)=$c_p$ for any ($1 \leq i \leq m$),
$ed$($S^i$, $S^j$)=$c_d$ for any ($1 \leq i  \leq m$, $1 \leq j \leq m$, $i \neq j$),
and $c_d= \theta *c_p$. According to Theorem 2 and the above assumptions, we have

\begin{eqnarray}
\nonumber 
{ed(C,O)}^2 &=& \frac1{m^2}[m\sum_{i=1}^m{ed(S^{i},O)}^2-\sum_{i=1}^{m-1}\sum_{j=i+1}^m{ed(S^{i},S^{j})}^2]  \\ \nonumber 
&=& {c_p}^2-\frac{m-1}{2m}{c_d}^2 \\ \nonumber 
&=& (1-\frac{m-1}{2m}{\theta}^2){c_p}^2 
\end{eqnarray}

Therefore,  

\begin{equation}
ed(C,O)=\sqrt{(1-\frac{m-1}{2m}{\theta}^2)}*c_p 
\end{equation}

By definition, $ed(C,O)$ cannot be negative 
and \((1-\frac{m-1}{2m}{\theta}^2) \geq 0\) should hold. Therefore, 
for a given $m$, $\theta$ must have a maximal limit.
If $m=2$ and $\theta=2$, then $ed(C,O)=0$. 
$\theta=2$ must be the maximal value in this case.
Likewise, when $m=3$, the maximal value for $\theta$ is $\sqrt{3}$.

\begin{figure}
\centerline{\psfig{figure=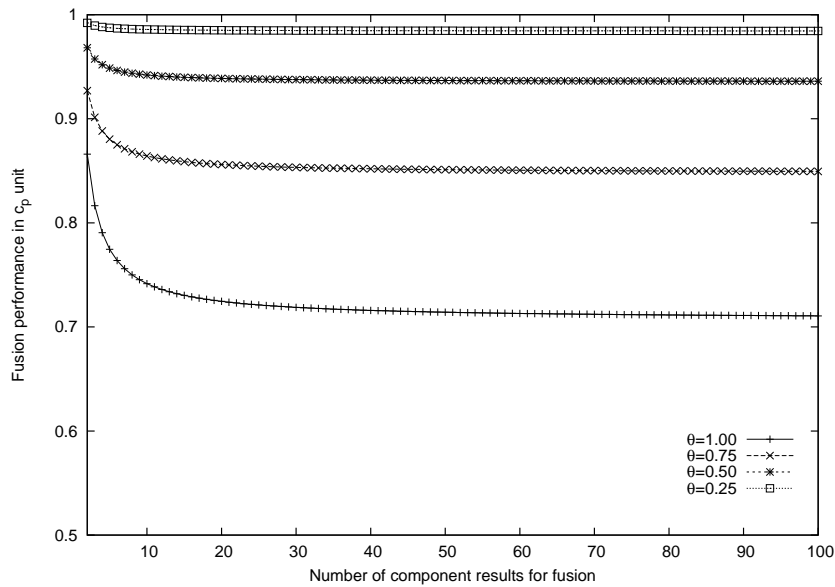,width=11.2cm}}
\caption{The impact of component classifier number on ensemble performance}
\end{figure}

Fig. 1 shows the values of $ed(C,O)$ in $c_p$ unit for $ \theta =0.25,0.5, 0.75, 1$ and $m=2, 3, \cdots ,100$.
From Fig. 1 we can see that in all four cases, $ed(C,O)$ decreases with $m$.
However, as $m$ becomes larger and larger, the rate of decrease becomes smaller and smaller. 
When $m$ tends to infinity, $ed(C,O)$ approaches $\sqrt{1-0.5{\theta}^2}$ $c_p$ units.
They are 0.984, 0.935, 0.848, and 0.707 when $ \theta $ equals to 0.25, 0.5, 0.75, and 1, respectively. 
However, adding more component results does not help much if there is already a large number.
For example, when $\theta=1.0$ and $m=27$, $ed(C,O)$ is 0.720 $c_p$ units.
It is close to the limit of 0.707 $c_p$ units. It suggests that fusing 30 or more 
component results may not be very useful for further improving ensemble performance. 
This has been observed in some empirical studies before, such as in \cite{OshiroPB12}, and others.


Theorem 1 tells us that the ensemble performance is at least as good as
the average performance of all the component classifiers involved. 
This may not be positive enough for many applications of the technique.
Theorem 3 indicates that it may take too much time to choose a subset from 
a large group of candidates for good ensemble performance.
This problem may be solved in other ways.
In particular, if we want the ensemble performance
to be better than the best component classifier, more favourable conditions
are required for those component classifiers. It means that we need to 
apply some restrictions to all the component classifiers involved.
Theorem 5 can be useful for this.

\noindent \textbf{Theorem 5.}
In a space $X$, $\mathcal{S}$ = $\{S^1$, $S^2$,$\cdots$, $S^m\}$ and $O$ are known points.
At least one of the points in $\mathcal{S}$ is different from the others.
$C$ is the centroid of $S^1$, $S^2$,$\cdots$, $S^m$.
If $ed(S^1,O)=ed(S^2,O)=\cdots=ed(S^m,O)$,
then $ed(C,O)<ed(S^1,O)$ must hold.

Proof: According to Theorem 2, we have

\begin{eqnarray}
\nonumber 
{ed(C,O)}^2&=&\frac{1}{m}\sum_{i=1}^m{ed(S^i,O)}^2-\frac{1}{m*m}\sum_{i=1}^{m-1}\sum_{j=i+1}^m{ed(S^i,S^j)}^2  \\ \nonumber   
&<&\frac{1}{m}\sum_{i=1}^m{ed(S^i,O)}^2={ed(S^1,O)}^2
\end{eqnarray}

Therefore, we obtain $ed(C,O)<ed(S^1,O)$. \hfill $\Box$

Theorem 5 tells us if all the component classifiers are equally effective,
then majority voting is able to do a better job than Theorem 1's guarantee.
In practice, this has been implemented in various situations.
For example, if using bagging with random forest or neural networks \citep{OshiroPB12,YangZZW13},
then we are can generate a large number of almost equally-effective component classifiers. 
Good performance is achievable by fusing such classifiers.
See Section 4 for more discussions.

\subsection{Weighted majority voting}

Weighted majority voting is a generalization of majority voting. 
It is more flexible than its counterpart because different weighting schemes 
can be de- fined. It might be believed that both are similar, 
and this is indeed true for the cases when weights across component classifiers 
are largely similar. However, for the weighted majority voting, 
we have little interest in the universe of all possible weighting schemes, 
but are rather focused on the optimum weighting scheme. 
We delve into the following two questions especially. 

\begin{enumerate}

\item How to find the optimum weights for a group of component classifiers? 

\item What are the properties of weighted majority voting with the optimum weights?

\end{enumerate}

Let us begin with the first question. 
In a space $X$, $\mathcal{S}$ = $\{S^1,S^2,\cdots,S^m\}$ and $O$ are known points.
Let $F$ be the fused point for linear combination of $m$ points in $\mathcal{S}$ with weighting 
$w^1,w^2,\cdots,w^m$.

\begin{equation}
ed(F,O)^2=\sum_{i=1}^{n}\sum_{j=1}^{p}{(\sum_{k=1}^{m}{(w^k*s_{ij}^{k})}-o_{ij})^2}
\end{equation}

Our goal is to minimize $ed(F,O)^2$.
Assuming $f(w^1,w^2,\cdots,w^m)=ed(F,O)^2$, we have
\[ \frac{\partial f}{\partial w^q}= \sum_{i=1}^{n}\sum_{j=1}^{p}2(\sum_{k=1}^{m}(w^k*s_{ij}^{k})-o_{ij})s_{ij}^q\]
Let \( \frac{\partial f}{\partial w^q}=0 \) for ($q$=1,2,$\cdots$,$m$), then 
\[\sum_{k=1}^{m}w^k\sum_{i=1}^{n}\sum_{j=1}^{p}{(s_{ij}^ks_{ij}^q)}=\sum_{i=1}^{n}\sum_{j=1}^{p}{o_{ij}s_{ij}^q} \]

Let \(a_{qk}=\sum_{i=1}^{n}\sum_{j=1}^{p}s_{ij}^{k}s_{ij}^{q} \) for $(1 \leq q \leq m)$ and $(1 \leq k \leq m)$, 
and \( b_q=\sum_{i=1}^{n}\sum_{j=1}^{p}o_{ij}s_{ij}^q \) for $(1 \leq q \leq m)$.
Thus we obtain the following $m$ linear equations with $m$ variables $w^1,w^2,\cdots,w^m$:

\begin{equation}
\label{eq:leastS}
 \left( \begin{array}{c@{\:+\:}c@{\:+\:}c@{\:+\:}c@{\:=\:}c}
a_{11}w^1 & a_{12}w^2 & \cdots & a_{1m}w^m & b_1 \\
a_{21}w^1 & a_{22}w^2 & \cdots & a_{2m}w^m & b_2 \\
\multicolumn{5}{c}{\dotfill}  \\
a_{m1}w^1 & a_{m2}w^2 & \cdots & a_{mm}w^m & b_m 
\end{array} \right) 
\end{equation}

The optimum weights can be calculated by 
finding the solution to these $m$ linear equations. 
Note that minimizing $ed(F,O)^2$ and minimizing $ed(F,O)$ is equivalent 
for us to find the optimum weights because $ed(F,O)$ can not be negative.

\noindent \textbf{Theorem 6.}
In a $n*p$ dimensional space $X$, $\mathcal{S}$=\{$S^1$, $S^2$,$\cdots$, $S^m$\} 
and $O$ are known points. If every point in $\mathcal{S}$ is linearly independent from the others, 
then the above process and Equation \ref{eq:leastS} can find the unique solution 
to the problem. 

Proof: The independency of each point in $\mathcal{S}$
indicates that $m \leq n*p$ holds. For the same reason, any point 
that can be represented linearly by these $m$ points has a unique representation. 
We may write $ed(F,O)$ as $f(w^1,w^2,\cdots,w^m)$, which is a continuous function.
In the whole space, there is only one minima and no other saddle points or maxima. 
The point at which all partial derivatives of the function to all variables equal to zero must be
the minimum point. The $m$ equations set up in Equation \ref{eq:leastS} 
are able to find the point with a unique representation of weights. \hfill $\Box$

Intuitively, in a $n*p$ dimensional space $X$,
$m$ points in $\mathcal{S}$=\{$S^1$, $S^2$,$\cdots$, $S^m$\} comprise a subspace $X'$ in $X$.  
For any point $O$, there exists one and only one point in $X'$ that has the shortest distance to $O$.
This point can be linearly represented by $S^1$, $S^2$,$\cdots$, $S^m$.

Theorem 6 can be explained as follows. 
For a (training) dataset with $n$ instances, $p$ classes, and $m$ classifiers,
each of the classifiers gives a score for each instance and each class.
For each instance, we also have real labels relating to all the classes.
Then we are able to find a group of weights 
$w^1,w^2,\cdots,w^m$ for {$S^1$, $S^2$,$\cdots$, $S^m$} 
to achieve the best ensemble performance by weighted majority voting.

The following Theorems 7 and 8 answer the second question.

\noindent \textbf{Theorem 7.} 
In a $n*p$ dimensional space $X$, $\mathcal{S}$ = \{$S_1$, $S_2$,$\cdots$, $S_m$\} is a group of points,
$m<n*p$ and all $m$ points are independent of each other. $O$ is an ideal point. 
For a given number $m'$ ($2 \leq m'<m$), the problem is to find a subset of $m'$ points from $\mathcal{S}$
so as to let the fused point of 
the chosen $m'$ points by weighed majority voting to the ideal point $O$ as close as possible. 
This problem is NP-hard.

Proof. As shown in Theorem 6, for each group of $m'$ points, their optimal weights can be 
calculated by least-squares with a time complexity of O(${m'}^2*n*p$) \citep{BoydV04}.
To choose $m'$ points from a total number of $m$ points, there are

\[ \binom{m}{m'}=\frac{m!}{m'!(m-m')!} \]

combinations. When $m'$ approaches $m/2$, the number of combinations grows exponentially  
with $m$. Note that because each group of points is different from the other groups, their weights need
to be calculated by least-squares separately. Therefore, the problem is an NP-hard problem. \hfill $\Box$

\noindent \textbf{Theorem 8.}
In a $n*p$ dimensional space $X$, $\mathcal{S}^1$=\{$S^1$, $S^2$,$\cdots$, $S^m$\},
$\mathcal{S}^2$=\{$S^1$, $S^2$,$\cdots$, $S^m$, $S^{m+1}$\},
and $O$ is an ideal point. If the optimum weights are used for both $\mathcal{S}^1$
and $\mathcal{S}^2$, then the performance of Group $\mathcal{S}^2$
is at least as effective as that of Group $\mathcal{S}^1$.

Proof. Assume that $w^1, w^2,\cdots,w^m$ are optimum weights for  
$S^1$, $S^2$,$\cdots$, $S^m$ of $\mathcal{S}^1$ to obtain 
the best performance. For $\mathcal{S}^2$, if using the same weights $w^1, w^2,\cdots,w^m$ 
for $S^1$, $S^2$,$\cdots$, $S^m$, and 0 for $S^{m+1}$,  
then the ensemble performance of $\mathcal{S}^2$ will be the same as that of $\mathcal{S}^1$. Note 
that the above weighting scheme is by no means the best for $\mathcal{S}^2$ and it is also 
possible to find more profitable weights for $\mathcal{S}^2$. \hfill $\Box$

Theorem 8 can be explained as follows: for a given dataset, consider 
two groups of classifiers. 
Group 1 has $n$ classifiers: $cf_1$,$cf_2$,$\cdots$,$cf_m$,
and Group 2 has $m+1$ classifiers, $cf_1$,$cf_2$,$\cdots$,$cf_m$,$cf_{m+1}$.
$m$ classifiers in both groups are the same.
If using weighted majority voting with the optimum weights, then 
the ensemble performance of Group 2 
is at least as effective as that of Group 1.

\noindent \textbf{Corollary 8.1.} 
In a $n*p$ dimensional space $X$, 
assume that weighted majority voting is applied with the optimum weights.
When more and more points are added,
the ensemble performance is monotonically non-decreasing. 

Proof. It can be proven by applying Theorem 8 repeatedly. \hfill $\Box$

Intuitively, when more and more points are added, the subspace becomes bigger and bigger.
During this process, it is possible to find new points that has the shortest distance 
to the ideal point.

Corollary 8.1 can be explained in this way. Assume for a given dataset, 
an ideal ensemble is implemented by weighted majority voting with the optimum weights. 
When more and more classifiers are added into such an ensemble,
its performance is non-decreasing monotonically.

Theorem 2 tells us about how individual classifier performance ($ed(S^i,O)$)
and dissimilarity between classifiers ($ed(S^i,S^j)$) affect ensemble performance
with the majority voting scheme.
We may regard weighted majority voting as a variation of majority voting.
Before applying Theorem 2, weighted majority voting changes the positions
of all component results' position by a linear weighting scheme, 
thus Equation \ref{eq:theorem2} becomes

\begin{equation}
\label{eq:theorem2w}
ed(C,O)=\frac1m\sqrt{m\sum_{i=1}^m{ed(w^i \cdot S^i,O)}^2-\sum_{i=1}^{m-1}\sum_{j=i+1}^m{ed(w^i \cdot S^i,w^j \cdot S^j)}^2}
\end{equation}

where $w^i \cdot S^i=\sum_{j=1}^{n}\sum_{k=1}^{p}w^i*s_{jk}^i$. After that, 
both can be treated in the same way.

\section{Discussion}

In Section 3 we have set up a geometric framework and presented the properties of 
majority voting and weighted majority voting.
Now we are in a good position to compare
the two different levels of geometric frameworks and 
answer the questions raised in the first section of this paper.

\subsection{Dataset-level vs. instance-level frameworks}

A major objective of the ensemble problem is to try to provide a solution 
for all the instances in a dataset. A framework at different levels has certain impact
on the way we can deal with the problem. For an instance-level framework, 
we need an approach to expand it to cover all the instances
in the whole dataset, while the dataset-level framework does not need it.

In the dataset-level framework, all the instances in the whole dataset are
concatenated to form a super-instance. Thus, all the properties stand 
in the instance-level framework also stand in the dataset-level framework.  

However, a few differences need to be noted.  
One is the dimensionality of the geometric space involved.
For a classification problem with $p$ classes and a dataset with $n$ instances,
the dimensionality of the instance-level geometric space is $p$,
while that of the dataset-level geometric space is $p*n$.

How to calculate optimal weights for weighted majority voting is another 
place where we may have different solutions. 
The solution given in Section 3 of this paper is to minimize the Euclidean distance between
the linear combination of all the component points and the ideal point (refer to Equation 11).
Recall that all the instances in the dataset is transformed to a single super-instance.
However, for the instance-level framework, we still need to consider 
multiple instances together. One possible way is to minimize the sum of the distance
over all instances, or

\begin{equation}
\sum_{i=1}^n{ed(F,O)}=\sum_{i=1}^{n}\sqrt{\sum_{j=1}^{p}{(\sum_{k=1}^{m}{(w^k*s_{ij}^{k})}-o_{ij})^2}}
\end{equation}

It is tricky to optimise Equation 14 directly. To simplify, 
we may optimise $\sum_{i=1}^n{ed(F,O)^2}$ instead \citep{Wu&Crestani15,BonabC18}. 
In this way, it is the same as Equation 11. 
It demonstrates that there are connections between the two levels of frameworks.
For the instance-level framework, optimising $\sum_{i=1}^n{ed(F,O)^2}$ approximates optimising   
$\sum_{i=1}^n{ed(F,O)}$. 
On the other hand, for the dataset-level framework, optimising $ed(F,O)^2$ is the same as 
optimising $ed(F,O)$. It means that the weights obtained are optimum. 

\subsection{The size of ensemble}

As discussed in Section 3, we proved that the number of 
component classifiers has positive impact on ensemble performance 
for both majority voting and weighted majority voting.
However, this contradicts the assertion in \cite{BonabC19}:
for a multi-classification problem with
$k$ classes, $k$ is the ideal number of base classifiers to constitute 
an optimum ensemble by weighted majority voting.  
Let us analyse this further.

\noindent \textbf{Example 6.} Consider a classification problem with two classes 
and three base classifiers. One instance is shown in 
the figure below. Weighted majority voting is used for combination.

\begin{center}
\begin{tikzpicture}

\draw [fill] (0,0) circle [radius=0.06];
\draw [fill] (1.2,-1.2) circle [radius=0.06];
\draw [fill] (-0.2,-1.2) circle [radius=0.06];
\draw [fill] (1.3,0.3) circle [radius=0.06];
\node [left] (s1) at (0,0) {$S^1$};
\node [right](s2) at (1.2,-1.2) {$S^2$};
\node [left](s3) at (-0.2,-1.2) {$S^3$};
\node [right](o) at (1.3,0.3) {$O$};

\draw (0.5,-0.5);
\node (p) at (0.5,-0.5){ };

\draw[dashed] (0,0) --  (1.2,-1.2);
\draw[dashed] (0.5,-0.5) --  (1.3,0.3);
	
\draw[solid] (0.7,-0.3)--(0.9,-0.5);
\draw[solid] (0.7,-0.7)--(0.9,-0.5);

\end{tikzpicture}
\end{center}

In the figure above, it shows a two-dimensional space with three points $S^1$, $S^2$, $S^3$
to be a combination. The ideal point is $O$.
We can see that combining three points can lead to the optimal results of zero distance than fusing any two.
Therefore, in this example the assertion in \cite{BonabC19} even does not hold at the instance level. 
However, as shown in this example, 
for a $n$ dimensional space, $n+1$ independent points are enough. \hfill $\Box$

We may add some restrictions to the points involved. For example, 
for a binary classification problem, we let all the points to be 
on the line segment of [0,1] and [1,0]. The ideal point is either [1,0] or [0,1].
In this way, a maximum of two points are needed
for the optimal fusion results. In the figure below, any two of the three points 
$S^1$, $S^2$, and $S^3$ are competent for this task.

\begin{center}
\begin{tikzpicture}

\draw [fill] (0,3) circle [radius=0.06];
\draw [fill] (3,0) circle [radius=0.06];
\draw [fill] (1,2) circle [radius=0.06];
\draw [fill] (1.7,1.3) circle [radius=0.06];
\draw [fill] (2.2,0.8) circle [radius=0.06];
\node [left] (o1) at (0,3) {$O^1$[0,1]};
\node [right](o2) at (3,0) {$O^2$[1,0]};
\node [left](s1) at (1,2) {$S^1$};
\node [left](s2) at (1.7,1.3) {$S^2$};
\node [left] (s3) at (2.2,0.8) {$S^3$};

\draw[dashed] (0,3) --  (3,0);

\end{tikzpicture}
\end{center}

Anyhow, the assertation at the instance level is not very useful.
To consider the problem in a more realistic way, we need to look at it at the dataset level.
A dataset usually comprises at least a good number of instances. 
If we consider three instances with a binary classification problem, then the 
dimensionality of the dataset-level geometric framework is up to 2*3=6, and not 2 any more.
If the dataset has more instances, then we may include even more independent classifiers.
With an increased number of base classifiers, 
we will likely get better ensemble performance (Corollary 8.1).

On the other hand, we can obtain the same conclusion as Corollary 8.1 
even under the instance-level framework.
Assume that the whole dataset has $n$ instances. $F_i$ and $O_i$ are the fused point and
the ideal point for instance $t_i$, respectively.
The optimal weighting for $m$ base classifiers are $w^1$,$w^2$,$\cdots$,$w^m$.
Then we have

\begin{equation}
\sum_{i=1}^n{ed(F_i,O_i)}=\sum_{i=1}^{n}\sqrt{\sum_{j=1}^{p}{(\sum_{k=1}^{m}{(w^k*s_{ij}^{k})}-o_{ij})^2}}
\end{equation}

Now one more base classifier is added. We can set a new weighting scheme as 
$w_{new}^1=w^1$,$w_{new}^2=w^2$,$\cdots$,$w_{new}^m=w^m$, $w_{new}^{m+1}=0$.

\begin{equation}
\sum_{i=1}^n{ed_{new}(F_i,O_i)}=\sum_{i=1}^{n}\sqrt{\sum_{j=1}^{p}{(\sum_{k=1}^{m+1}{(w_{new}^k*s_{ij}^{k})}-o_{ij})^2}}
\end{equation}

Then $\sum_{i=1}^n{ed(F_i,O_i)}=\sum_{i=1}^n{ed_{new}(F_i,O_i)}$.
$w^1$,$w^2$,$\cdots$,$w^m$ is the optimal weighting scheme for $m$ base classifiers, while 
$w_{new}^1$,$w_{new}^2$,$\cdots$,$w_{new}^m$, $w_{new}^{m+1}$ may not be optimal
for $m+1$ base classifiers.
Therefore, if the optimal weighting scheme is used, then fusing
$m+1$ base classifiers can achieve at least the same performance as fusing
$m$ base classifiers. This is exactly what Corollary 8.1 tells us.

\subsection{Answer to some questions}

In Section 1, we listed some outstanding questions. Now let us discuss them one by one.
 
Question 1: What is the difference between majority voting and weighted majority voting?

In a sense, both weighting schemes can potentially enhance ensemble performance. 
However, there are certain aspects, including their abilities, to consider. 
Weighted majority voting can be better than the best component classifier 
if optimum weights are used, while majority voting can be better than 
the average of all component classifiers. Majority voting is a ``mild'' method 
because all the component results are treated equally and the centroid is the solution, 
while weighted majority voting is an ``extreme'' method 
because it does not treat all component results equally 
and it takes the most effective solution from all possible ones.

Question 2: When should we use majority voting rather than weighted majority voting, or vice versa?

A general answer is: in cases majority voting does not
work well, then weighted majority voting should be used.
Now a further question is: when is majority voting a good method?
Performance of all component results, dissimilarity of all pairs
of component results, number of component results have positive impact on ensemble 
performance. A judicious decision should consider these factors thoroughly.
More specifically, Theorem 2 answers this question quantitatively. 
One easy noticeable situation is that when all the component classifiers 
are of equal or very close performance, then  
majority voting may be able to achieve better ensemble performance 
than the best component classifier (Theorem 5).

Question 3: There are a lot of weighting assignment methods for weighted majority voting,
which one is the best?

The least squares is the best weighting assignment method
for the measure of Euclidean distance. Compared with 
many others, it is efficient and effective at the same time. 
Almost all other weighting assignment methods are either heuristic
or optimisation methods. For the former, its effectiveness is not guaranteed;
for the latter, it is timing-consuming.

Question 4: How does the number of component classifiers affect ensemble performance?

The number of component classifiers has a positive effect on ensemble performance. 
The situation is straightforward for weighted majority voting. 
When more and more component results are added to an ensemble,
its performance becomes better and better. 
However, the situation for majority voting is more complicated. 
Adding  more component classifiers into an ensemble cannot always improve performance. 
When the number is small, then its impact on ensemble performance is large. 
When the number increases, its impact becomes smaller.

Question 5: How does each component classifier affect ensemble performance?

It is related to the first question. For majority voting,
each contributes equally; for weighted majority voting, each contributes differently 
in order to get the optimal results for the whole data set.
If a very good component classifier is added, then weighted majority voting
can take advantage of it. On the other hand, for majority voting,
if many component classifiers are poor, then 
a few good ones will not be able to improve performance very much.

Question 6: How does performance of component classifiers affect ensemble performance?

Performance of component classifiers is the most important aspect that
affect ensemble performance.
For majority voting, a high-performance point is able to move the centroid of the group 
closer to the ideal point. 
For weighted majority voting, a high-performance point very likely enables 
the subspace to expand with some points closer to the ideal point.

Question 7: How does diversity of component classifiers affect ensemble performance?

For diversity, there are many different types of definitions before.
In this paper, we define it as the dissimilarity distribution of all pairs of component results. 
Apart from performance, diversity is another aspect that impacts ensemble performance significantly.
For majority voting, a comparative investigation about it and performance has been done in Subsection 3.1. 
Based on a simplified situation, the importance ratio between diversity and performance is 
calculated to be in the range of (0.25,0.5], varying with the number of component classifiers. 
For weighted majority voting, high diversity among component results will 
make the subspace bigger, thus it is more likely to find closer points to the ideal point 
in such a space.

One final comment about the framework is: 
all the theorems in the geometric framework hold when the Euclidean distance
is used for measuring performance. When other metrics are used, 
the conclusions we obtain may hold for many of the instances, but not every single instance.
However, there is strong correlations between any other meaningful performance metrics and
the Euclidean distance. If enough instances are observed, we may expect 
consistent conclusions.

\section{Empirical investigation}

In this section we are going to investigate how theoretical conclusions presented in 
Section 3 can be confirmed for practical use.
Within the geometric framework, all the theorems hold perfectly when Euclidean distance is used
as performance metric on the same dataset.
We would like to see how they behave when the conditions are partially satisfied.
Specifically, two points are considered:

\begin{itemize}

\item Usually classification accuracy or some other metrics, rather than Euclidean distance,
is used for performance evaluation, although Euclidean distance and all those commonly used metrics 
are strongly correlated;

\item When the component classifiers and ensemble models are trained using some training data,
they need to be used and tested in the test dataset, which may be somewhat different
from the training dataset.

\end{itemize}

For the above purpose, we carried out the empirical investigation by using the WEKA machine learning suite\footnote{http://www.cs.waikato.ac.nz/ml/weka/} 
and 20 datasets downloaded from the UCI Machine 
Learning Repository\footnote{ https://archive.ics.uci.edu/ml/index.php}. 
The main statistics of the these 20 datasets are listed in Table 2.

\begin{table}
\begin{center}
\caption{Statistics of two datasets used in the study}
\begin{tabular}{lrrr} \hline

Data set & No. of instances  &  No. of attributes & No. of classes \\ \hline
Anneal&898&39&6\\
Credit-g&1000&21&2\\
Data-cortex-nuclear&1080&81&8\\ 
Germancredit &1000&21&2\\ 
Hypothyroid&3772&30&4\\ 
Kr-vs-kp&3196&37&2\\
Mfeat-factors&2000&217&10\\ 
Mfeat-fourier&2000&77&10\\ 
Mfeat-karhunen&2000&65&10\\ 
Mfeat-zernike&2000&48&10\\ 
Optdigits&5620&65&10\\ 
Pendigits&10992&17&10\\ 
Secom&1567&591&2\\ 
Segment&2310&20&7\\ 
Semeion&1593&257&10\\ 
Sick&3772&30&2\\ 
Soybean&683&36&19\\ 
Spambase&4601&58&2\\ 
Splice&3190&61&3\\ 
Vowel&990&14&11\\ \hline

\end{tabular}
\end{center}
\end{table}

We set up ensembles (random forests) by using up to 30 random trees as base classifiers.
Both classical random forest and weighted random forest are tested. The weighting scheme for weighted
random forest is to optimise the Euclidean distance. 
For a given dataset, all the instances are divided into two disjoint partitions after stratification, 
in which 80\% is taken as the training partition and the remaining 20\% as the test partition.
During the training process, sampling with replacement is used to extract data 
for training classifiers. 30 component random trees are generated with $K$ randomly chosen features,
where $K$ is the square root of the total number of features in that dataset. 
First we use all 30 trees for an ensemble.
Then one of them is chosen randomly and removed, and the remaining 29 are ensembled.
The above process is repeated until there are two, which is the minimal number for an ensemble.

Both random forest and weighted random forest are tested.
Apart from these two combination schemes, the best component classifier and average of all
component classifiers are also calculated for comparison.
Two metrics including Euclidean distance and classification accuracy are 
used for performance evaluation. Tables 3 and 4 show the results for the training partition
and test partition, respectively. Each figure in these two tables is the average of 30 ensembles.

\begin{table}
\begin{center}
\caption{Classification performance of classical random forest and 
its weighted counterpart in the training partition; a pair of figures in bold indicates
that random forest performs better than its weighted counterpart for the given measure,
and a pair of underlined figures indicates there is a tie between the two methods.}
\begin{tabular}{|l|rr|rr|} \hline

& \multicolumn{2}{|c|}{Random forest} & \multicolumn{2}{|c|}{Weighted random forest} \\ 
\cline{2-5}
\raisebox{1.5ex}[0pt]{Dataset} & Accuracy(\%) & Distance & Accuracy(\%) & Distance \\ \hline

Anneal&99.93&1.74&99.94&1.42\\
Credit-g&\underline{98.74}&7.03&\underline{98.74}&6.63\\
Data-cortex-nuclear&99.72&3.90&99.73&3.35\\ 
Germancredit&98.72&7.05&98.73&6.65\\ 
Hypothyroid&\textbf{99.91}&3.64&\textbf{99.88}&2.88\\ 
Kr-vs-kp&99.86&4.25&99.87&3.89\\
Mfeat-factors&99.58&6.39&99.60&5.97\\ 
Mfeat-fourier&98.88&9.83&98.96&8.96\\ 
Mfeat-karhunen&99.23&8.67&99.30&7.75\\ 
Mfeat-zernike&98.38&10.73&98.50&10.00\\ 
Optdigits&99.66&9.54&99.68&9.01\\ 
Pendigits&99.87&7.90&99.88&7.72\\ 
Secom&99.13&5.46&99.16&5.30\\
Segment&99.84&3.98&99.86&3.75\\ 
Semeion&99.18&\underline{7.37}&99.19&\underline{7.37}\\
Sick&\underline{99.89}&4.15&\underline{99.89}&3.87\\ 
Soybean&99.47&4.47&99.52&4.25\\ 
Spambase&99.60&7.90&99.61&7.66\\ 
Splice&99.39&10.17&99.41&9.14\\ 
Vowel&99.39&5.51&99.42&4.95\\ \hline

\end{tabular}
\end{center}
\end{table}

\begin{table}
\begin{center}
\caption{Classification performance of classical random forest and its weighted counterpart 
in the test partition; a pair of figures in bold indicates
that random forest performs better than its weighted counterpart for the given measure,
and a pair of underlined figures indicates there is a tie between the two methods.}

\begin{tabular}{|l|rr|rr|} \hline

& \multicolumn{2}{|c|}{Random forest} & \multicolumn{2}{|c|}{Weighted random forest} \\ 
\cline{2-5}
\raisebox{1.5ex}[0pt]{Dataset} & Accuracy(\%) & Distance & Accuracy(\%) & Distance \\ \hline

Anneal&\textbf{98.96}&2.15&\textbf{98.95}&2.08\\
Credit-g&72.39&\textbf{8.56}&72.73&\textbf{8.63}\\
Data-cortex-nuclear&98.95&3.72&98.97&3.29\\ 
Germancredit&73.19&8.49&73.36&8.56\\ 
Hypothyroid&98.92&3.90&99.04&3.47\\ 
Kr-vs-kp&98.74&4.34&98.90&4.10\\
Mfeat-factors&95.73&6.38&95.84&6.18\\ 
Mfeat-fourier&\textbf{80.88}&11.27&\textbf{80.66}&11.10\\ 
Mfeat-karhunen&92.74&8.88&92.78&8.43\\ 
Mfeat-zernike&75.68&\textbf{11.92}&75.71&\textbf{11.94}\\ 
Optdigits&96.03&10.66&96.09&10.37\\ 
Pendigits&\underline{98.68}&8.16&\underline{98.68}&8.09\\
Secom&92.82&\textbf{6.41}&93.02&\textbf{6.45}\\ 
Segment&\textbf{97.33}&4.60&\textbf{97.29}&4.57\\ 
Semeion&85.06&9.47&85.76&9.17\\ 
Sick&98.63&4.22&98.65&4.04\\ 
Soybean&93.05&4.47&93.34&4.25\\ 
Spambase&94.27&\textbf{8.96}&94.30&\textbf{9.00}\\ 
Splice&91.57&11.36&92.02&10.70\\ 
Vowel&94.59&5.67&94.60&5.42\\ \hline

\end{tabular}
\end{center}
\end{table}

\begin{table}
\begin{center}
\caption{Statistics of paired T test and correlation test for the results of classical random 
forest and weighted random forest}

\begin{tabular}{|l|l|rr|rr|} \hline

& & \multicolumn{2}{|c|}{Correlations} & \multicolumn{2}{|c|}{Compare means} \\ 
\cline{3-6}
\raisebox{1.5ex}[0pt]{Partition} & \raisebox{1.5ex}[0pt]{Metric} & Correlation & Significance & Mean & Significance \\ \hline

Training & Accuracy & .998 & .000 & -.025 & .003 \\ 
Training & Distance & .995 & .000 &  .458 & .000 \\ 
Testing  & Accuracy &1.000 & .000 & -.124 & .012 \\ 
Testing  & Distance & .998 & .000 &  .188 & .001 \\ \hline

\end{tabular}
\end{center}
\end{table}

\begin{figure*}
\begin{center}

\par
\centerline{\hbox{ 
\psfig{figure=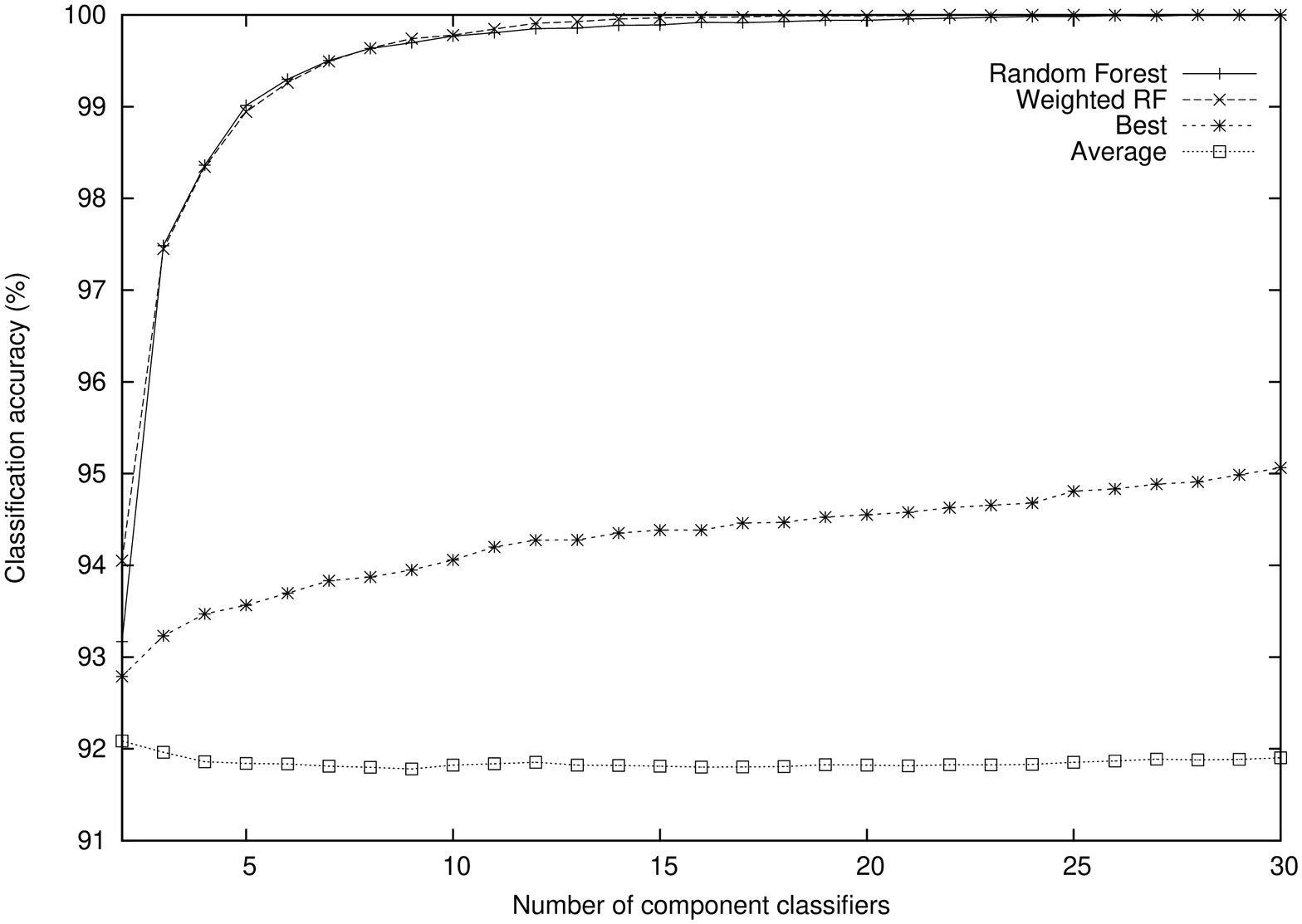,width=5cm,angle=270}
\psfig{figure=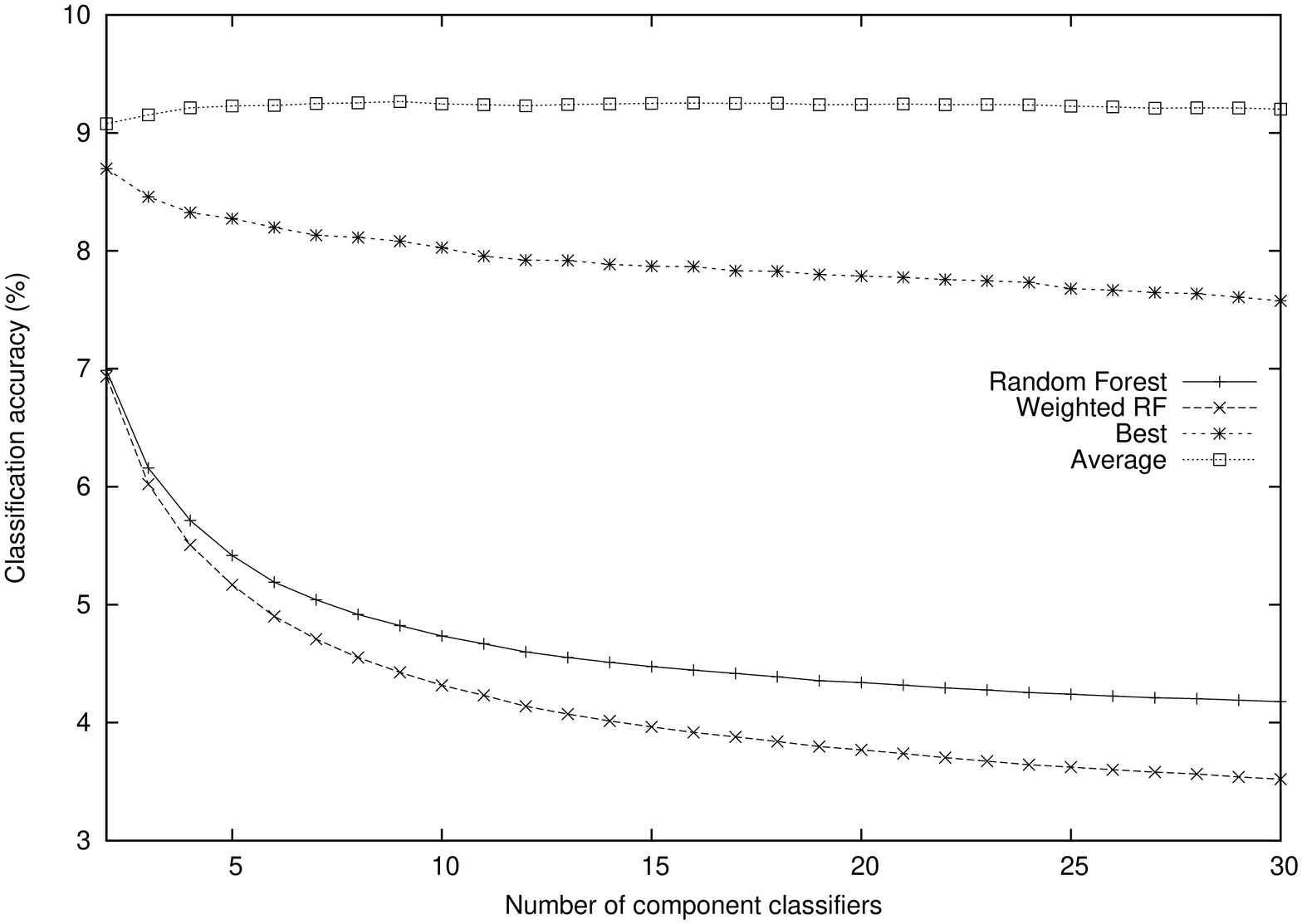,width=5cm,angle=270}
}}
\vspace{2mm}
(a) Training part/Accuracy \hspace{15 mm} (b) Training part/Euclidean distance
\par
\par
\centerline{\hbox{ 
\psfig{figure=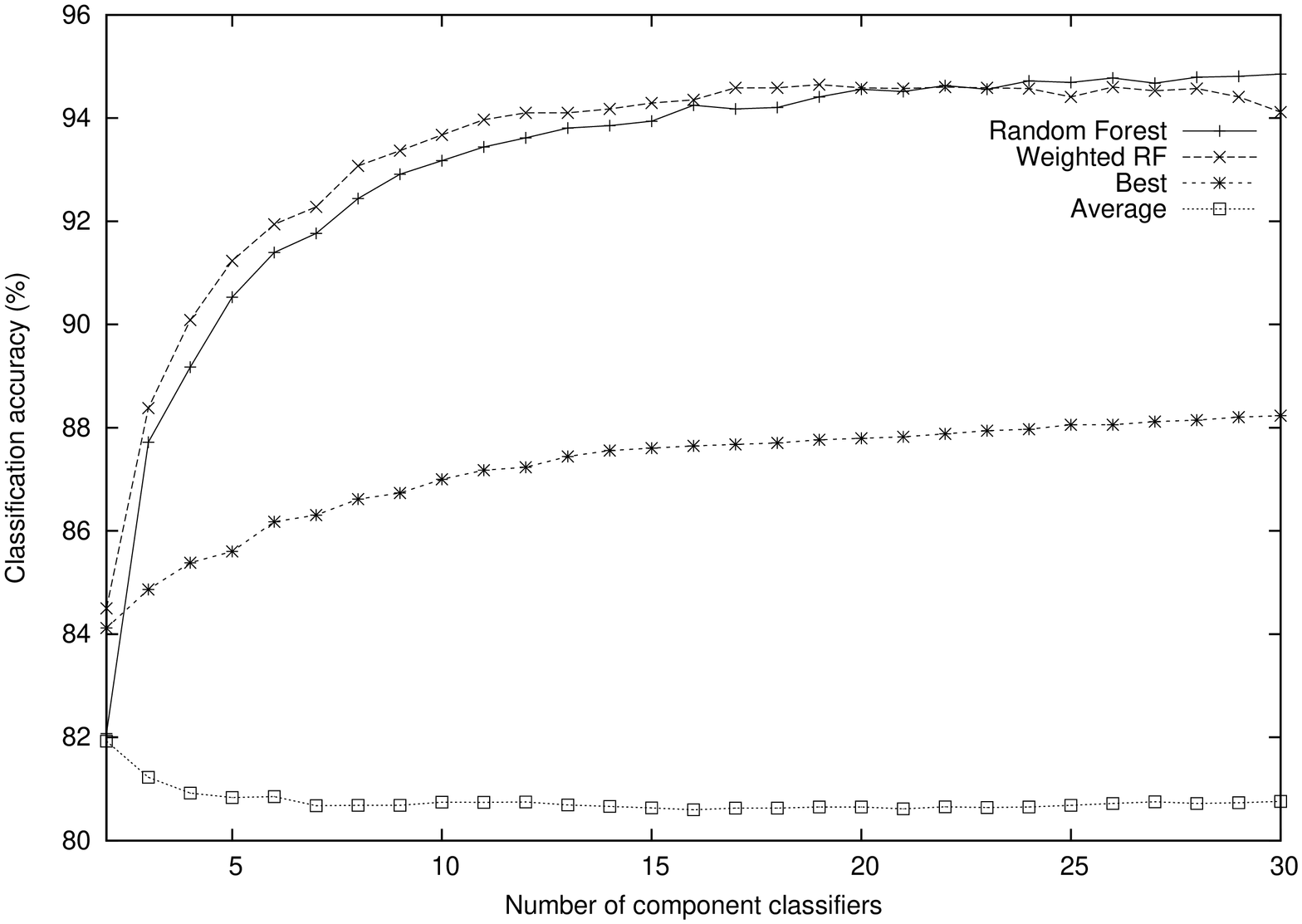,width=5cm,angle=270}
\psfig{figure=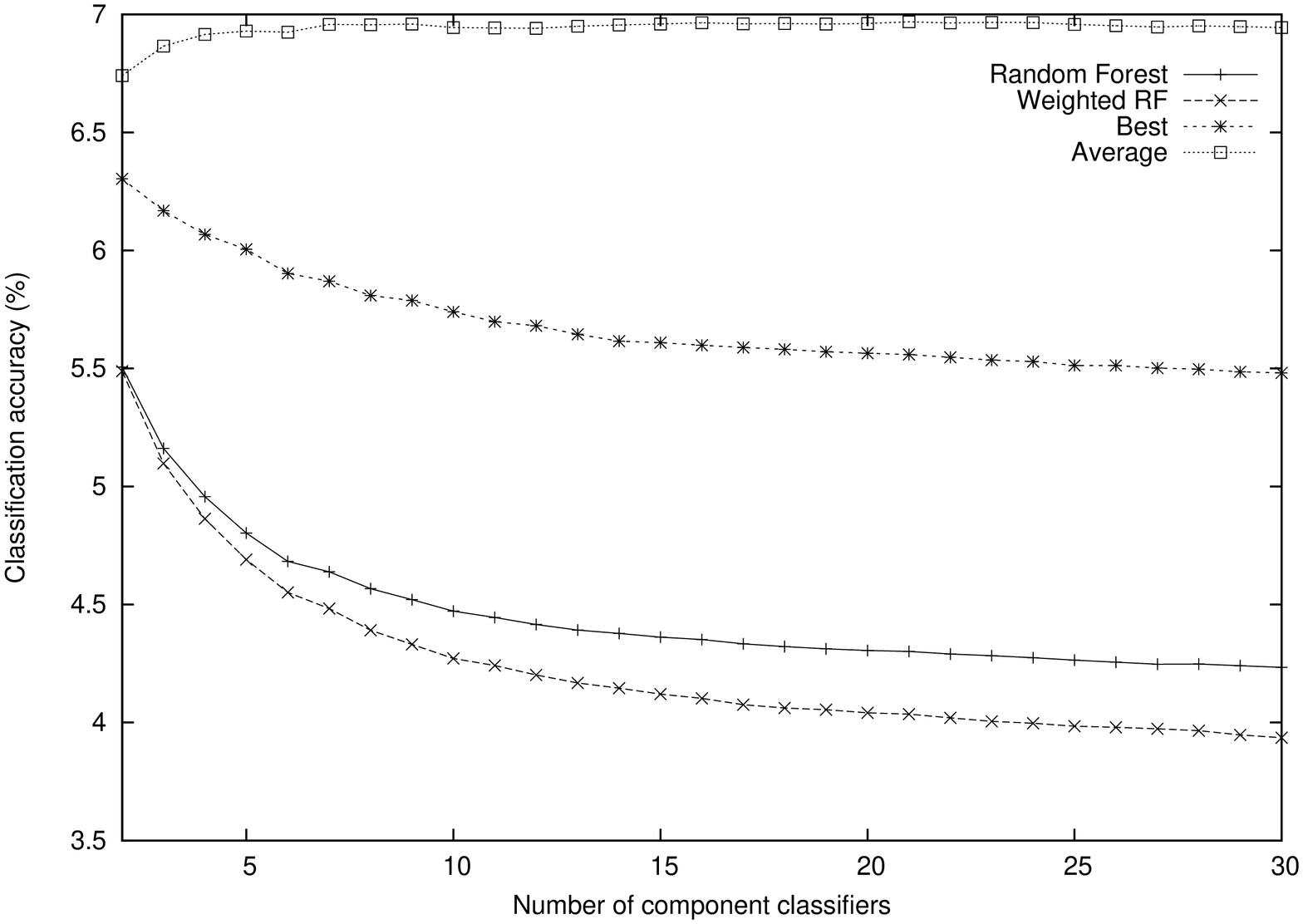,width=5cm,angle=270}
}}
\vspace{2mm}
(c) Test part/Accuracy \hspace{15 mm} (d) Test part/Euclidean distance
\par
\caption{Performance comparison of two combination schemes with 
different number of base classifiers for dataset Soybean (``Best'' denotes the best base
classifier and ``Average'' denotes the average of all base classifiers involved.)}

\end{center}
\end{figure*}

For the training partition, we can see from Table 3 that the weighted random forest is always better than
the classical random forest when Euclidean distance is used as the metric for evaluation. The only exception
is Semeion, in which there is a tie. This is trivial because Euclidean distance is the optimisation goal.
When accuracy is used as the metric, the classical random forest is better than the weighted counterpart 
on one dataset and there are ties on two datasets, while the weighted random forest is better than the classical 
on other 17 datasets.
As Table 4 shows that the situation is very similar for the test partition,  
although the classical random forest is a little better than it is in the training partition.
Considering all the datasets, it is better than the weighted 
on three and four datasets, for accuracy and Euclidean distance, respectively. 
Besides, there is a tie on one dataset when measured by accuracy. 
However, for the majority datasets, the weighted random forest performs better than 
the classical.

Although the weighted random forest performs better than the classical one 
on both training and test partitions, it is noticeable that the difference between them is very small.
This is because all generated base classifiers are very close in performance and 
it is a very good condition for the classical random forest to do good work.
To further compare the difference between them, we carried out paired $t$ test plus
correlation test for them. Table 5 shows the result.
From Table 5 we can see that the difference between classical random forest
and weighted random forest is significant in both partitions, either accuracy or 
Euclidean distance is used as the metric. Among them, the lowest significance level, .012,
is still very high and happens in the test partition with accuracy as the metric.
On the other hand, in all four cases, the correlation between two groups of results
are always very strong. All four correlation coefficients are very close to 1.

Finally, we take a look at the effect of number of base classifiers on ensemble performance.
Figure 2 shows the performance of different ensemble methods with the Soybean dataset.
Performances of random forest, weighted random forest, the best base classifier, and 
the average of all base classifiers are presented. Note that the curves for all other 
datasets are very similar in shapes. It is clear that performance increases 
for both random forest and weighted random forest when more 
base classifiers are involved. From Figure 2, we have a few other observations. 
First, for the metric of accuracy, performance of both random forest and weighted random forest
increases very rapidly with the number of base classifiers when the number is small.
When the number is bigger, it slows down very quickly. When the number is 10 or more, 
the increase rate becomes very small. On the other hand, such a phenomenon is not so
prominent for the Euclidean distance. Second, the difference between 
random forest and weighted random forest is more noticeable for the Euclidean distance than
for accuracy. It seems that only a small percentage of the deduction in distance
has been transferred to the increase in accuracy.

From the above experimental results, it demonstrates that the Euclidean distance 
is a very useful measure for performance evaluation and especially for optimisation. 
Therefore, we conclude that in general, the theorems we obtain from the geometric
framework still make sense even when other metrics such as accuracy is used for performance evaluation.

\section{Conclusions}

In this paper, we have presented a dataset-level geometric framework for 
ensemble classifiers. The most important advantage of the framework is 
it makes ensemble learning a deterministic problem.
Both performance and dissimilarity can be measured by the same metric - the Euclidean distance,
thus it is a good platform for us to understand 
the fundamental properties of ensembles clearly and investigate many issues in ensemble classifiers,
such as the impact of multiple aspects on ensemble performance, 
predicting ensemble performance, selecting a small number of base classifiers 
to get efficient and effective ensembles, etc.
Otherwise, it is very challenging to grasp even an incomplete picture. 
This is why up to now some of the properties of majority voting 
and weighted majority voting have not been fully understood.

Compared with the instance-level framework in \cite{Wu&Crestani15,BonabC19}, 
the dataset-level framework presented in this paper is a step forward.
It maps the ensemble classifier problem for a whole dataset into one multi-dimensional
space, thus it is more convenient for us to investigate the properties of ensembles.
Otherwise, we have to deal with multiple spaces at the same time, each for one instance.
To find out the collective properties in those spaces is more complicated.
Based on the dataset-level framework, we have deduced some useful theorems which had not been found before.

An empirical investigation has also been conducted to see
how those theorems in the geometric framework hold when accuracy rather than the Euclidean distance
is used for performance evaluation. 
The experimental results show that the theorems are still meaningful for other metrics.

In this paper, the setting for the proposed framework is traditionally with a batch of training data.
In recent years, data stream classification has attracted some attention \citep{GomesBEB17}.
How to adapt the geometric framework for this is worth further research.
Especially, incorporating dynamic updates is a key point.
Another research topic is multi-model data fusion \citep{GaoLCZ20}. Again how to adapt the framework to
support multimodal data fusion is an interesting research issue.
One possible solution is to use a separate framework for each mode and then to combine them. 
These research issues remain to be our future work.


\end{document}